\documentclass[10pt,twocolumn,letterpaper]{article}

\usepackage[pagenumbers]{cvpr} 

\usepackage{graphicx}
\usepackage{multirow}
\usepackage{amsmath}
\usepackage{amssymb}
\usepackage{booktabs}
\usepackage[bottom]{footmisc}
\usepackage{caption}
\usepackage{subcaption}
\usepackage[T1]{fontenc} 

\usepackage[font=small,skip=0pt]{caption}
\usepackage{amssymb}
\usepackage{pifont}
\newcommand{\cmark}{\ding{51}}%
\newcommand{\xmark}{\ding{55}}%

\usepackage[pagebackref,breaklinks,colorlinks]{hyperref}

\usepackage[capitalize]{cleveref}
\crefname{section}{Sec.}{Secs.}
\Crefname{section}{Section}{Sections}
\Crefname{table}{Table}{Tables}
\crefname{table}{Tab.}{Tabs.}

\def\cvprPaperID{7} 
\def\confName{CVPR}
\def\confYear{2022}

\begin{document}

\title{Human Body Shape Classification Based on a Single Image}

\author{Cameron Trotter\thanks{Research undertaken whilst on internship at Levi Strauss \& Co.}\\
Newcastle University, UK\\
{\tt\small c.trotter2@ncl.ac.uk}
\and
Filipa Peleja\\
Levi Strauss \& Co., Belgium\\
{\tt\small filipapeleja@gmail.com}
\and
Dario Dotti\\
Levi Strauss \& Co., Belgium\\
{\tt\small ddotti@levi.com}
\and
Alberto de Santos\\
Levi Strauss \& Co., Belgium\\
{\tt\small adesantos@levi.com}}

\maketitle

\begin{abstract}
    There is high demand for online fashion recommender systems that incorporate the needs of the consumer's body shape. As such, we present a methodology to classify human body shape from a single image. This is achieved through the use of instance segmentation and keypoint estimation models, trained only on open-source benchmarking datasets. The system is capable of performing in noisy environments owing to to robust background subtraction. The proposed methodology does not require 3D body recreation as a result of classification based on estimated keypoints, nor requires historical information about a user to operate - calculating all required measurements at the point of use. We evaluate our methodology both qualitatively against existing body shape classifiers and quantitatively against a novel dataset of images, which we provide for use to the community. The resultant body shape classification can be utilised in a variety of downstream tasks, such as input to size and fit recommendation or virtual try-on systems. 
\end{abstract}

\section{Introduction}
\label{sec:intro}

When inside a physical store, it is possible for customers to see and feel products they wish to purchase in real time. With online channels this is not possible. Instead customers are required to purchase products before they have interacted with them - only being able to do so once the product arrives. For online fashion retailers this can be problematic, as customers purchasing a product must make size and fit decisions based solely on guides and intuition. This can lead to a negative experience for the customer if a product does not fit as they wished, resulting in potential emotional impact \cite{kim_relationship_2010} and ultimately in them returning the purchased product. 

Studies have shown that size and fit inaccuracies are a major factor in dissatisfaction with clothing \cite{otieno_fit_2007, alexander_clothing_2005, sindicich_assessment_2011}. A high return rate, up to 30\% in the online fashion retail industry by some estimates \cite{de_trade-offs_2016}, can also cause issues for the retailer who must manage the returns process. Furthermore there are environmental concerns due to increased transportation of products \cite{velazquez_environmental_2019, cullinane_bricks_2009, wells_retail_2019}. Because of these factors it is extremely important for online fashion retailers to aid the customer as much as possible through the buying process. This is achieved currently through personalised purchase recommendations, based primarily on past purchase and return data.

\begin{figure*}
	\begin{center}
		\includegraphics[width=\linewidth]{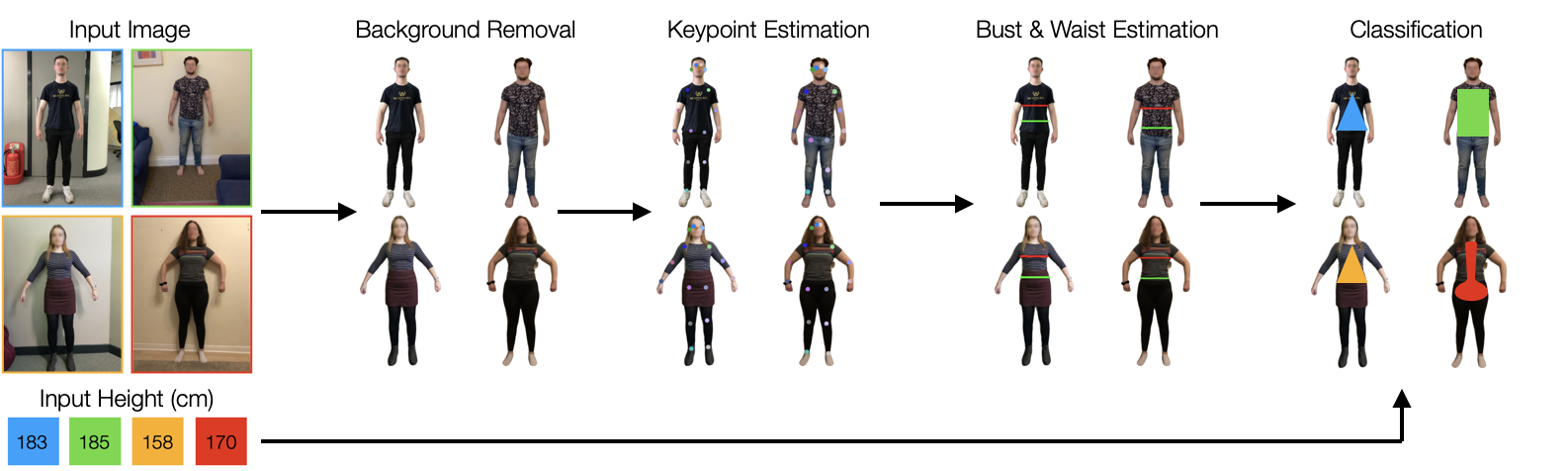}
	\end{center}
	\caption{A high level overview of data flow through the proposed system.}
	\label{fig:pipeline}
\end{figure*}

Body shape is an important factor when selecting clothing, greatly impacting how the product will fit the wearer \cite{sattar_fashion_2019}. Different aspects of a clothing product will make it more suitable for certain body shapes, helping compliment or draw attention from certain parts of the wearer's body. However body shape is often not considered as a factor for automated fashion recommendation. The use of 3D body scanners to gather accurate measurement data is infeasible when selling via online channels, as this would require customers to be in possession of the required hardware. In contrast, ownership of cameras capable of standard image capture is near ubiquitous thanks to the popularity of smartphones. A significant amount of research has been undertaken to recreate 3D body shape from a single image \cite{alldieck_tex2shape_2019, pavlakos_expressive_2019, zhu_detailed_2019, zanfir_thundr_2021}, however these recreations are prone to errors such as misplaced limbs \cite{bogo_keep_2016}.

Rather than aiming to produce a full accurate 3D recreation of the customer’s body, this paper shows that body shape can instead be classified into one of multiple classes utilising more readily available sources of data: a single image and user height. This is achieved using a pipeline of deep learning models trained on open-source benchmarking datasets for instance segmentation and keypoint estimation, post-processing the outputs and utilising these to classify body shape. The methodology proposed does not require previous historical data to operate, avoiding the common cold start problem \cite{lika_facing_2014, schein_methods_2002}, and produces a classification which can then be used to aid fashion recommendations downstream.

\section{Related Work}\label{sec:relatedWork}

Recent years have seen a marked increase in the use of machine learning in the consumer fashion space. For use in online retail specifically, a large corpus of work focuses on recommending products or styles to customers similar to those they have previously purchased. This is achieved via textual attributes \cite{kucukbay_supervised_2019, sekozawa_one--one_2011, stefani_cfrs_2019}, image data \cite{nicosia_image-based_2019}, or a combination of both \cite{stan_intelligent_2019, zhou_fashion_2019, bracher_fashion_2016, liu_magic_2013, agarwal_personalizing_2018}.

Size recommendation is also an active area of research, both to handle vanity sizing issues \cite{aydinoglu_imagining_2012} and imperfections in the manufacturing process leading to deviations in expected and actual garment measurements \cite{donmezer_measurements_2021}. Approaches mostly focus on embedding generation \cite{sheikh_deep_2019, misra_decomposing_2018, abdulla_size_2017, heinz_lstm-based_2017}, although alternative approaches do exist \cite{sembium_recommending_2017, karessli_sizenet_2019, dokoohaki_attention_2021, lasserre_meta-learning_2020}.

One major disadvantage of the approaches to product and size recommendation highlighted is they do not utilise body shape, missing one of the most important aspects dictating a customer's purchase -- fit. Vitali \textit{et al.} aim to acquire customer body measurements to improve the fit of tailored products using a multi-camera set up \cite{vitali_acquisition_2018}, however this is infeasible to roll out at scale.

Utilising standard images and using these to recreate a 3D representation of the body has gained prominence in recent years. Systems such as Tex2Shape \cite{alldieck_tex2shape_2019}, THUNDR \cite{zanfir_thundr_2021}, and work presented by Pavlaoks \textit{et al.} \cite{pavlakos_expressive_2019} and Zhu \textit{et al.} \cite{zhu_detailed_2019} aim to estimate 3D body shape from a single image. However these models can fail under certain conditions, such as misplacing limbs or incorrectly estimating the shape of certain body parts. Nonetheless they have been used for fashion-oriented body shape estimators in the past, with Neophytou \textit{et al.} proposing ShapeMate which utilises a PCA model trained on human 3D scan data to recreate body shape, classifying into one of nine classes \cite{neophytou_shapemate_2013}. 

Other work has shown that if the goal is to classify the body's shape rather than generate a full recreation, it is possible to achieve this without the need for a 3D conversion. SmartFit proposed by Foysal \textit{et al.} \cite{foysal_smartfit_2021} aims to classify body type by processing images through an edge detection algorithm. Edges are then passed through either a SURF \cite{bay_speeded-up_2008} feature extractor and K-Nearest Neighbours clustering algorithm or a Convolutional Neural Network to classify body shape. However, it is not clear how well the system would perform with noisy backgrounds which may be common with customer-uploaded images. Hidayati \textit{et al.} utilise unsupervised learning to cluster humans into body shape classifications based on multiple textual attributes to learn the relationship between body shape and compatibility of clothing styles \cite{hidayati_what_2018}.

Based on a review of the above work and the importance of fit in the online fashion retail space, it is clear there exists a need for a system which can estimate a user's body shape without the use of specialised hardware, does not require large volumes of historical data as a starting point, and is capable of operating when the user is in a noisy environment. Work proposed in this paper fulfills these requirements. By utilising only a single image captured by a standard camera, customers are not required to own specialised hardware. By operating only on the information provided by the user at the time, no historical data-points are required. By utilising a state of the art segmentation model and a robust post-processing methodology the system is capable of operating on images which contain noisy background environments.

\begin{table*}
    \centering
    \begin{tabular}{|c|c|c|c|c|}
        \hline
        \textbf{Paper}                                     & \textbf{\# Classes} & \begin{tabular}[c]{@{}c@{}}\textbf{Full Set of}\\\textbf{User Measurements}\end{tabular} & \textbf{Requires 3D Recreation} & \textbf{Background Noise Removal}  \\ \hline
        ShapeMate \cite{neophytou_shapemate_2013}          & 9                   & \xmark                      & \cmark                  & \xmark                        \\ \hline
        SmartFit \cite{foysal_smartfit_2021}               & 4                   & \xmark                      & \xmark                  & \xmark                        \\ \hline
        Hidayati \textit{et al.} \cite{hidayati_what_2018} & 7                   & \cmark                      & \xmark                  & \xmark                        \\ \hline
        Ours                                               & 5                   & \xmark                      & \xmark                  & \cmark  \\ \hline                     
        \end{tabular}
        \caption{Qualitative comparison of proposed methodology against related work.}
        \label{tab:paperEval}
\end{table*}

\subsection{Comparison Against Related Work}\label{subsec:EvalOnRelated}

An evaluation of our proposed methodology against the most closely related work was conducted, mainly Hidayati \textit{et al.} \cite{hidayati_what_2018}, ShapeMate \cite{neophytou_shapemate_2013}, and Smartfit \cite{foysal_smartfit_2021}. As the data used in these papers is not readily available\footnote{SmartFit: Unable to acquire dataset from link provided in paper. Correspondence author contacted but no reply received.} a qualitative evaluation was undertaken, a summary of which is provided in Table \ref{tab:paperEval}.

Unlike Hidayati \textit{et al.} our method requires only height measurement as input, reducing the required workload of users. Measurements are not required for ShapeMate however this method stipulates the use of a 3D recreation of the user's body which is computationally more expensive than our proposed methodology. Furthermore, noise caused by background can effect classifications obtained from both ShapeMate and SmartFit; this effect is reduced in our system through the use of instance segmentation for background removal. 

\section{Methodology}\label{sec:Methodology}

The methodology proposed to determine a body shape utilises a pipeline of deep learning models alongside a body shape classification module. Models present in the system pipeline are trained on community-standard open-source benchmarking datasets. An overview of the pipeline's structure can be seen in Figure \ref{fig:pipeline}. 

The pipeline takes in two inputs: an image of a human standing in a neutral anatomical pose, and a numerical value representing their height. Height information is required to allow for accurate distance measurements. The system uses this \textit{in lieu} of requiring the user to stand with an object of known size for calibration purposes, which may be difficult to achieve given the aim of the system to be capable of use anywhere and at any time.

The input image is first passed through a background subtraction module. This is achieved using a DeepLabV3 \cite{chen_rethinking_2017} model with a ResNet101 backbone \cite{he_deep_2015}, trained on the PASCALVOC dataset \cite{everingham_pascal_2010} for instance segmentation. The model outputs a multi-class segmentation mask, post-processed into a binary mask denoting \texttt{human} or \texttt{background}, and is utilised to subtract the background from the input image. Background subtraction is required to reduce noise, increasing the precision of the bust, waist, and hip measurement estimations downstream.

The resultant background-subtracted image is then fed to an HRNet \cite{sun_deep_2019} model trained on the COCO 2017 Keypoints dataset\footnote{COCO 2017 Keypoints: \href{https://cocodataset.org/\#keypoints-2017}{cocodataset.org/\#keypoints-2017}} with the goal of keypoint estimation. This model outputs the estimated locations for multiple different body part pairs, allowing for accurate body shape classification.

By utilising the estimated shoulder and hip keypoints, the segmented human body, as well as \textit{a priori} knowledge of human anatomy, it is possible to estimate the locations of the human's bust and waistline. To the best of our knowledge no publicly available dataset for training keypoint estimators includes bust and waistline location ground-truths, however thanks to the proportionality of humans the training of HRNet on a dataset which contains these two keypoints in the training data is not required to gain an accurate and robust location estimation.

In order to utilise the calculated keypoints for body shape classification, the distances between keypoint pairs must be calculated. This is achieved by calculating a pixel-to-centimetre conversion, utilising the centimetre height input value and the pixel height of the segmentation mask.

\begin{figure}
	\begin{center}
        \includegraphics[width=\linewidth]{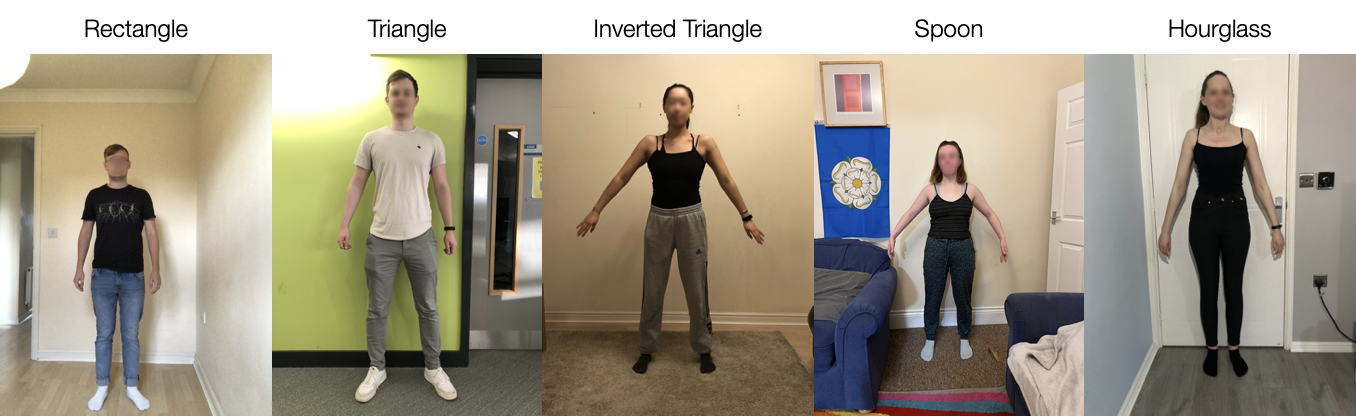}
	\end{center}
	\caption{Example images from the dataset and their body shape ground truth classifications. Images are captured of users in a natural anatomical pose and in noisy background environments.}
	\label{fig:classes}
\end{figure}

The calculated bust, waist, and hip locations as well as centimetre distances are fed into a classifier to determine the body shape. This classifier outputs one of five possible classes, based on work by Yim Lee \textit{et al.} \cite{yim_lee_comparison_2007}: \texttt{Rectangle}, \texttt{Triangle}, \texttt{Inverted Triangle}, \texttt{Spoon}, and \texttt{Hourglass}. Examples of the body types are shown in Figure \ref{fig:classes}.

The body shape classification can later be incorporated into product and size recommender systems, for example by passing as input to an embedding model, or utilised to improve clothing overlay in virtual try-on systems such as those proposed by Viadurre \textit{et al.} \cite{vidaurre_fully_2020} or Choi \textit{et al.} \cite{choi_viton-hd_2021}. 

\section{Evaluation}\label{sec:Evaluation}

In order to evaluate the system, image and measurement data from multiple participants was collected. In the images, participants were required to wear non-baggy clothing in a neutral anatomical pose. Measurement of their height in centimetres was recorded alongside bust, waist, and hip distance measurements.

\begin{figure}
	\begin{center}
        \includegraphics[width=0.8\linewidth]{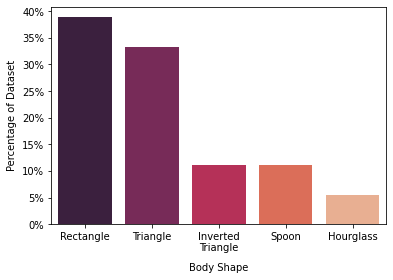}
	\end{center}
	\caption{Evaluation dataset class distribution.}
	\label{fig:dist}
\end{figure}

This evaluation dataset, which we provide access to\footnote{Dataset access via: \href{https://doi.org/10.25405/data.ncl.c.5875730}{doi.org/10.25405/data.ncl.c.5875730}.}, contains a 57:43\% split of male to female bodies. Body shape classes are imbalanced, indicative of the natural distribution of human body shapes. In Figure \ref{fig:dist} the percentage count of examples per class is highlighted. Classes \texttt{Rectangle} and \texttt{Triangle} are heavily represented in the dataset, believed to be caused by the natural lack of variation in male hip measurements \cite{molarius1999waist}. 

The proposed system achieves 71.4\% body shape accuracy on the evaluation dataset, a commendable score given the fine-grained nature of the problem. We achieve comparable accuracy to both ShapeMate \cite{neophytou_shapemate_2013} and SmartFit \cite{foysal_smartfit_2021}, however our system does not require the use of 3D recreation and is capable of operating in noisy environments. It is important to note however that due to data availability issues it is not possible to evaluate our system against the data used in these works. 

Whilst the system is able to accurately classify example images of the well represented body shapes, the overall accuracy metric is reduced due to incorrect classifications of the \texttt{Hourglass} class. Incorrect classifications of this shape are a result of high waist measurement error, an issue also present in ShapeMate. 

\begin{table}
    \centering
    \begin{tabular}{|c|c|c|} 
    \hline
    \multirow{3}{*}{\textbf{Measurement}} & \multicolumn{2}{c|}{\textbf{Absolute Average Error (cm)}}                                                                                   \\ 
    \cline{2-3}
                                          & \begin{tabular}[c]{@{}c@{}}\small{\textbf{Correct~}}\\\small{\textbf{Classifications}}\end{tabular} & \begin{tabular}[c]{@{}c@{}}\small{\textbf{Incorrect~}}\\\small{\textbf{Classifications}}\end{tabular}  \\ 
    \hline
    Bust                                  & $2.29 \pm 1.73$                                                                               & $6.51 \pm 4.69$                                                    \\ 
    \hline
    Waist                                 & $3.95 \pm 3.03$                                                                                & $5.08 \pm 3.90$                                                    \\ 
    \hline
    Hip                                   & $2.85 \pm 1.72$                                                                                & $5.29 \pm 4.48$                                                    \\
    \hline
    \end{tabular}
    \caption{Absolute average measurement error with standard deviations for correct and incorrect classifications.}
    \label{tab:measurementError}
\end{table}

Table \ref{tab:measurementError} shows the absolute average measurement error in centimetres between correct and incorrect classifications. As can be expected, for correct body shape classifications the system produces low measurement error. For incorrect classifications the error is higher, which may be attributed to perspective distortion due to camera angle. Misclassifications can broadly be attributed to one of two issues: lower than expected hip keypoint placement, and incomplete background subtraction around the bust or waist. Our proposed methodology for body part measurement achieves comparable results to those presented by ShapeMate \cite{neophytou_shapemate_2013} without the need to for a 3D recreation of the human to be generated.

Work proposed by Yim Lee \textit{et al.} \cite{yim_lee_comparison_2007} forms the basis for this work's body shape classifier. The formulae presented by Yim Lee \textit{et al.} are calculated based upon female measurements, however we observe no negative effect on body shape classification accuracy utilising these for males. Evaluating the methodology only on male dataset images gives an 83\% body shape accuracy. This may be helped by the absence of variation in male hip measurements causing a lack of class diversity by reducing the likelihood of males with a \texttt{Spoon} or \texttt{Hourglass} body shape \cite{molarius1999waist}. 

\begin{figure}
	\begin{center}
        \includegraphics[width=0.9\linewidth]{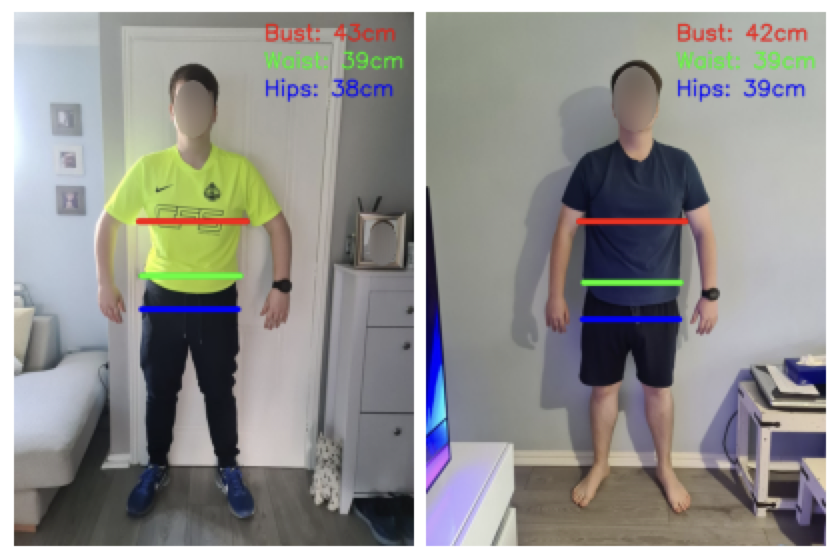}
	\end{center}
	\caption{The same participant in separate environments and clothing with estimated measurements and locations for bust, waist, and hips displayed.}
	\label{fig:bgClothes}
\end{figure}

Regardless of background environment or clothing, the proposed system should be robust enough to produce a consistent classification. To evaluate this, multiple images of the same participant standing in different locations and wearing different clothing were collected, an example of which can be seen in Figure \ref{fig:bgClothes}. Estimated measurements and locations for bust, waist, and hips have been overlaid. As can be observed, the system produces near identical location and measurement estimations in both images, resulting in correct \texttt{Inverted Triangle} classifications for the user in both cases. This highlights the versatility of the methodology proposed. 

\section{Conclusion}\label{sec:Conclusion}

This work presents a novel methodology for human body shape classification through a single image via a pipeline of deep learning models trained on common open-source benchmarking datasets. The system is capable of operating in noisy environments and without any required prior knowledge or the need to recreate the body in 3D space. The resultant classification can be useful in areas such as size and fit recommendation, as well as virtual try-on. It is hoped this work, as well as the release of the evaluation dataset, will spur subsequent developments in body shape classification without the need for 3D recreation.

\subsection{Future Work}\label{subsec:FutureWork}

Future work will focus on adapting the system to obtain more accurate body measurements. Currently the pixel-to-centimetre constant calculation does not take into account perspective distortion. This could be rectified by identifying an additional distortion term or through the application of automated image transformations. Furthermore, work will examine the inclusion of depth information through side-view imagery or video. Finally, work will continue on expanding the dataset to increase the number of available examples present for under-represented classes, as well as to include a wider distribution of body shapes, ages, and ethnic groups in order to better understand the system's robustness against a wider range of real world data.

{\small
\bibliographystyle{ieee_fullname}
\bibliography{references}
}

\end{document}